\documentclass[conference]{IEEEtran}
\IEEEoverridecommandlockouts

\usepackage{graphicx}
\usepackage{amsmath, amssymb}
\usepackage{float}
\usepackage{caption}
\usepackage{subcaption}
\usepackage{booktabs}
\usepackage{hyperref}
\usepackage{multirow}
\usepackage{array}
\usepackage{color}
\usepackage{tikz}
\usepackage{pgfplots}
\usepackage{cite}
\usepackage{bm}
\usepackage{adjustbox}
\usepackage[linesnumbered,ruled,vlined]{algorithm2e}
\usepackage{pifont}
\usepackage{siunitx}
\usepackage{mathtools}
\usepackage{url}

\def\BibTeX{{\rm B\kern-.05em{\sc i\kern-.025em b}\kern-.08em
    T\kern-.1667em\lower.7ex\hbox{E}\kern-.125emX}}

\usetikzlibrary{shapes.geometric, arrows.meta, positioning, calc, fit}

\hypersetup{
    colorlinks=true,
    linkcolor={blue},
    citecolor={blue},
    urlcolor={blue}
}

\begin{document}

% Title
\title{DeepEyeNet: Adaptive Genetic Bayesian Algorithm Based Hybrid ConvNeXtTiny Framework For Multi-Feature Glaucoma Eye Diagnosis}

\author{\IEEEauthorblockN{Angshuman Roy$^{1}$,
                          Anuvab Sen$^{2}$, 
                          Soumyajit Gupta$^{3}$,
                          Soham Haldar$^{4}$,\\
                          Subhrajit Deb$^{5}$,
                          Taraka Nithin Vankala$^{6}$ and 
                          Arkapravo Das$^{7}$}
\IEEEauthorblockA{$^{1,3,4,5,6,7}$Indian Institute of Engineering Science and Technology, Shibpur, Howrah 711103, India\\$^{2}$Georgia Institute of Technology, Atlanta, GA 30332, USA\\
Email: angshuman1roy@gmail.com$^{1}$, 
asen74@gatech.edu$^{2}$,
gupta.soumyajit02@gmail.com$^{3}$,\\
sohamhaldar04@gmail.com$^{4}$,
subhrajitd28@gmail.com$^{5}$,
nithintaraka.v@gmail.com$^{6}$
and arkapravodas03@gmail.com$^{7}$} 
\vspace{-1 cm}}

\maketitle

\begin{abstract}
Glaucoma is a leading cause of irreversible blindness worldwide, emphasizing the critical need for early detection and intervention. In this paper, we present \textit{DeepEyeNet}, a novel and comprehensive framework for automated glaucoma detection using retinal fundus images. Our approach integrates advanced image standardization through dynamic thresholding, precise optic disc and cup segmentation via a U-Net model, and comprehensive feature extraction encompassing anatomical and texture-based features. We employ a customized ConvNeXtTiny-based Convolutional Neural Network (CNN) classifier, optimized using our \textit{Adaptive Genetic Bayesian Optimization (AGBO)} algorithm. This proposed AGBO algorithm balances exploration and exploitation in hyperparameter tuning, leading to significant performance improvements. Experimental results on the EyePACS-AIROGS-light-V2 dataset demonstrate that \textit{DeepEyeNet} achieves a high classification accuracy of 95.84\%, which was possible due to the effective optimization provided by the novel AGBO algorithm, outperforming existing methods. The integration of sophisticated image processing techniques, deep learning, and optimized hyperparameter tuning through our proposed AGBO algorithm positions \textit{DeepEyeNet} as a promising tool for early glaucoma detection in clinical settings.

\end{abstract}


\section{Introduction}
Glaucoma is a group of progressive optic neuropathies characterized by the degeneration of retinal ganglion cells and their axons, leading to structural damage to the optic nerve head and visual field defects \cite{weinreb2014pathophysiology}. It is one of the leading causes of irreversible blindness globally, affecting millions of people every year. Early detection and management are vital, as they can significantly slow disease progression and preserve vision \cite{weinreb2014pathophysiology}.

Traditional diagnostic methods, such as tonometry, perimetry, and gonioscopy, require specialized equipment and expertise, limiting their accessibility in resource-constrained settings. Fundus imaging provides a non-invasive and cost-effective alternative, enabling visualization of the retina and optic nerve head. Structural changes associated with glaucoma, such as increased cup-to-disc ratio (CDR) and neuroretinal rim (NRR) thinning, can be assessed from fundus images \cite{jonas2004optic}.

However, manual analysis of fundus images is time-consuming and subjective. Automated methods leveraging deep learning have shown promise in improving detection rates and consistency \cite{li2018efficacy, asaoka2019using, de2018clinically, JD}. Existing approaches often face challenges due to variations in image quality, illumination, and anatomical differences among patients \cite{gulshan2016development, raghavendra2018deep}. In this paper, we introduce \textit{DeepEyeNet}, a comprehensive framework that addresses these challenges through:

\begin{itemize}
    \item \textbf{Advanced Image Standardization}: Utilizing dynamic thresholding to normalize fundus images, enhancing the focus on regions of interest.
    \item \textbf{Precise Segmentation}: Employing a U-Net-based model for accurate segmentation of the optic disc and cup \cite{ronneberger2015unet}.
    \item \textbf{Comprehensive Feature Extraction}: Extracting anatomical and texture-based features, including novel metrics related to the NRR and blood vessels.
    \item \textbf{Optimized Classification}: Customizing a ConvNeXtTiny-based CNN classifier, optimized using a \textit{hybrid Adaptive Genetic Bayesian Optimization (AGBO)} algorithm.
\end{itemize}

Our contributions include integrating these components into a novel unified framework, introducing a hybrid AGBO algorithm for efficient exploration and exploitation, demonstrating superior performance over existing methods, and providing a practical solution for automated glaucoma detection.

\section{Proposed Methodology}

\subsection{Dataset and Preprocessing}

% Placeholder for dataset images: RG and NRG
\begin{figure}[H]
    \centering
    \begin{subfigure}{0.48\linewidth}
        \centering
        \includegraphics[width=0.9\linewidth, height = 3.42 cm]{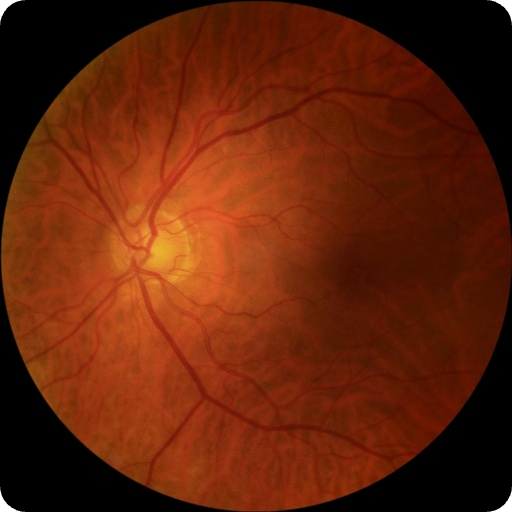}
        \caption{RG Image}
        \label{fig:RG_image}
    \end{subfigure}
    \hfill
    \begin{subfigure}{0.48\linewidth}
        \centering
        \includegraphics[width=0.9\linewidth, height = 3.42 cm]{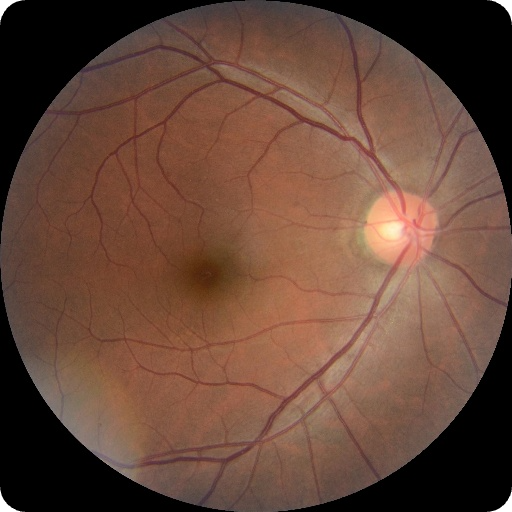}
        \caption{NRG Image}
        \label{fig:NRG_image}
    \end{subfigure}
    \caption{Sample images from the EyePACS-AIROGS-light-V2 dataset showcasing RG and NRG.}
    \label{fig:dataset_sample}
\end{figure}

We utilized the EyePACS-AIROGS-light-V2 dataset \cite{eyepacs2023dataset}, comprising 4,000 training images and 385 validation and test images for each class (referable and non-referable glaucoma).

\begin{figure}[H]
    \centering
    \begin{subfigure}{0.45\linewidth}
        \centering
        \includegraphics[width=\linewidth, height = 3cm]{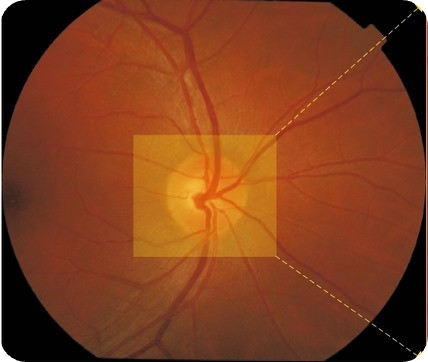}
        \caption{Normal Eye: Optic Cup, Disc, and NRR}
        \label{fig:normal_a}
    \end{subfigure}
    \hfill
    \begin{subfigure}{0.45\linewidth}
        \centering
        \includegraphics[width=\linewidth, height = 3cm]{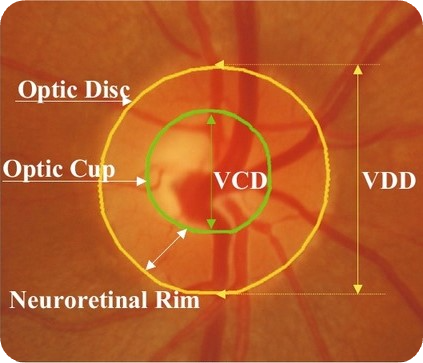}
        \caption{Normal Eye: Detailed Optic Structures}
        \label{fig:normal_b}
    \end{subfigure}
    
    \begin{subfigure}{0.45\linewidth}
        \centering
        \includegraphics[width=\linewidth, height = 3cm]{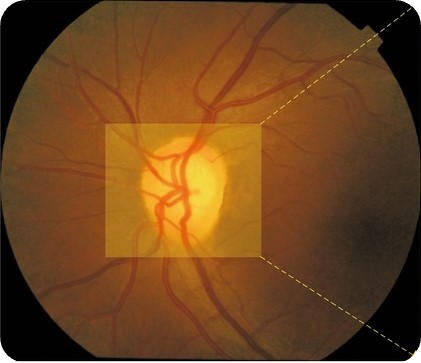}
        \caption{Glaucomatous Eye: Enlarged Optic Cup and Disc}
        \label{fig:glaucoma_c}
    \end{subfigure}
    \hfill
    \begin{subfigure}{0.45\linewidth}
        \centering
        \includegraphics[width=\linewidth, height = 3cm]{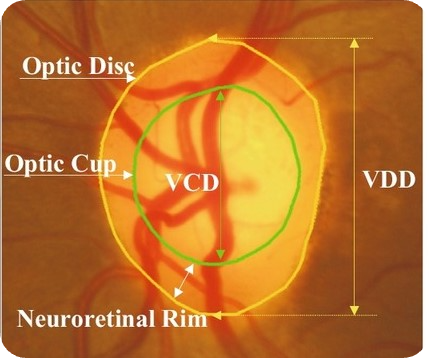}
        \caption{Glaucomatous Eye: Narrowed NRR}
        \label{fig:glaucoma_d}
    \end{subfigure}
    
    \caption{Fundus images illustrating normal and glaucomatous eyes with emphasis on key optic structures.}
    \label{fig:fundus_images}
\end{figure}

\subsubsection{Fundus Image Standardization}
To address variations in illumination and focus on the optic nerve head, we implemented dynamic thresholding for image standardization. The process includes grayscale conversion, histogram analysis, and adaptive thresholding using Otsu's method \cite{otsu1979threshold}. We then estimate the center and radius of the fundus image to crop and resize it to a standard size of $512 \times 512$ pixels.

% The following algorithm provides a step-by-step procedure for the segmentation and standardization of fundus images, which is crucial for maintaining consistency in image analysis and facilitating downstream processing tasks.

\begin{algorithm}[htbp]
\caption{Algorithm for Fundus Image Standardization}
\label{alg:standardization}
\SetAlgoLined
\KwIn{Input image $I_{\text{RGB}}$}

\textbf{Compute the following steps:}\\
\textbf{Step 1: Convert to Grayscale} \\
$ I_{\text{gray}} \leftarrow \text{ConvertToGrayscale}(I_{\text{RGB}}) $ \\
\textbf{Step 2: Compute Histogram} \\
$ H \leftarrow \text{ComputeHistogram}(I_{\text{gray}}) $ \\
\textbf{Step 3: Determine Threshold using Otsu's Method} \\
$ T \leftarrow \text{OtsuThreshold}(H) $ \\
\textbf{Step 4: Create Binary Mask} \\
$ M \leftarrow \text{CreateBinaryMask}(I_{\text{gray}}, T) $ \\
\textbf{Step 5: Estimate Center and Radius from Mask} \\
$ (x_c, y_c, r) \leftarrow \text{EstimateCenterRadius}(M) $ \\
\textbf{Step 6: Crop and Resize Around Center} \\
$ I_{\text{crop}} \leftarrow \text{CropAndResize}(I_{\text{RGB}}, (x_c, y_c), 512 \times 512) $ \\
\textbf{Step 7: Apply Circular Mask} \\
$ I_{\text{std}} \leftarrow \text{ApplyCircularMask}(I_{\text{crop}}, (x_c, y_c), r) $ \\
\KwOut{Standardized image $I_{\text{std}}$}
\Indp
\Return $I_{\text{std}}$
\end{algorithm}

This algorithm outlines a step-by-step procedure for segmenting and standardizing fundus images, ensuring consistent image analysis and facilitating downstream processing by producing a normalized standardized output.

% The algorithm concludes with a standardized output image that can be effectively used for consistent analysis, ensuring that the preprocessing steps have normalized the images for subsequent analysis.

% prev Fig.2

% \begin{figure}[H]
%     \centering
%     \includegraphics[width=\linewidth]{Pipeline.png}
%     \caption{Workflow of the proposed \textit{DeepEyeNet} framework.}
%     \label{fig:workflow}
% \end{figure}

\begin{figure*}[htbp]
    \centering
    \includegraphics[width=\textwidth]{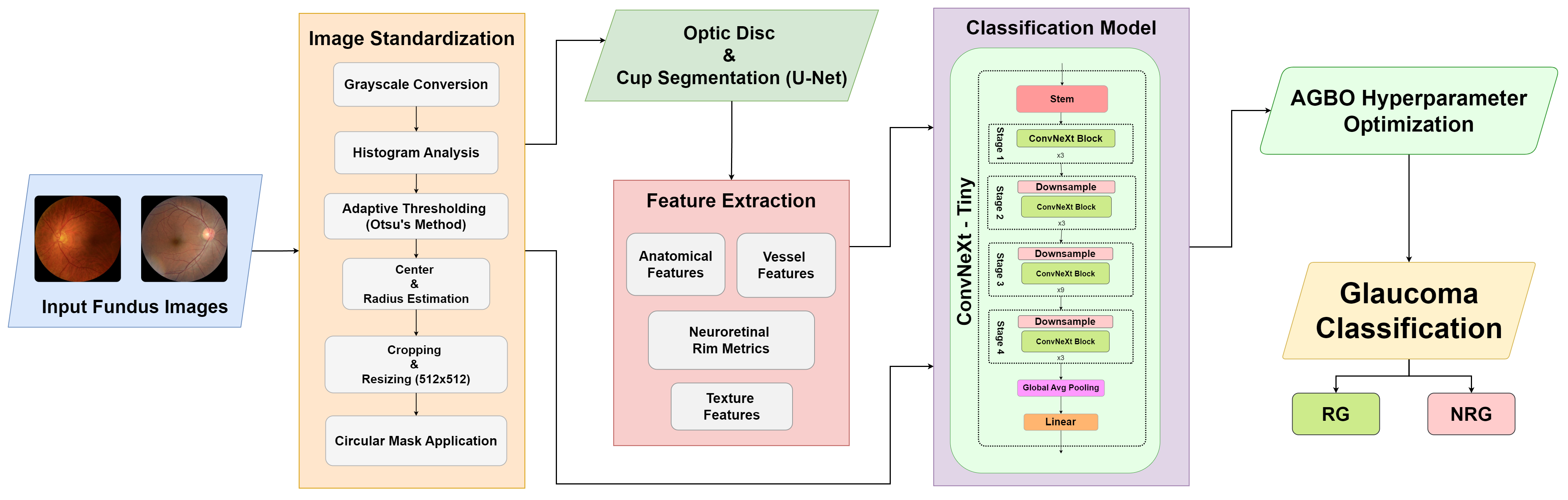}
    \caption{Workflow of the proposed \textit{DeepEyeNet} framework.}
    \label{fig:workflow}
\end{figure*}

\subsection{Overall Framework}
The \textit{DeepEyeNet} framework comprises four main components:

\begin{enumerate}
    \item \textbf{Image Standardization}
    \item \textbf{Optic Disc and Cup Segmentation}
    \item \textbf{Feature Extraction}
    \item \textbf{Classification}
\end{enumerate}
The \textit{DeepEyeNet} framework is designed to comprehensively analyze fundus images for medical diagnosis. The initial step, image standardization, ensures consistent quality and appearance of images. Optic disc and cup segmentation isolates key anatomical features, enabling accurate assessment of ocular health. Feature extraction derives essential visual features, which are then used to detect and predict ocular conditions like glaucoma or diabetic retinopathy\cite{Retinopathy}.

\subsection{Optic Disc and Cup Segmentation}
We employ a U-Net architecture \cite{ronneberger2015unet} for segmenting the optic disc and cup regions from the standardized images. The segmentation model is trained using a combination of Dice loss and cross-entropy loss to handle class imbalance and improve boundary delineation.

\begin{equation}
\mathcal{L}_{\text{seg}} = \lambda_{\text{Dice}} \mathcal{L}_{\text{Dice}} + \lambda_{\text{CE}} \mathcal{L}_{\text{CE}}
\end{equation}

\begin{equation}
\mathcal{L}_{\text{Dice}} = 1 - \frac{2 \sum_{i} p_i g_i}{\sum_{i} p_i + \sum_{i} g_i}
\end{equation}

where $p_i$ and $g_i$ are the predicted and ground truth labels, respectively.

\subsection{Feature Extraction}
From the segmentation results, we extract a set of features critical for glaucoma detection. The following metrics provide important information about the structural health of the optic nerve and can be used to detect early signs of glaucomatous damage.

\subsubsection{Anatomical Features}
The listed metrics provide key quantitative features of the optic disc and cup, such as areas and cup-to-disc ratios, which are crucial for assessing optic nerve health and detecting potential glaucoma.
\begin{itemize}
    \item \textbf{Disc Area} ($A_{\text{disc}}$)
    \item \textbf{Cup Area} ($A_{\text{cup}}$)
    \item \textbf{Cup-to-Disc Area Ratio} ($\text{CDR}_{\text{area}}$):

    \begin{equation}
    \text{CDR}_{\text{area}} = \frac{A_{\text{cup}}}{A_{\text{disc}}}
    \end{equation}

    \item \textbf{Vertical and Horizontal CDRs}:

    \begin{equation}
    \text{CDR}_{\text{vertical}} = \frac{H_{\text{cup}}}{H_{\text{disc}}}, \quad \text{CDR}_{\text{horizontal}} = \frac{W_{\text{cup}}}{W_{\text{disc}}}
    \end{equation}

    where $H$ and $W$ represent the height and width, respectively.
\end{itemize}

\subsubsection{Neuroretinal Rim Metrics}

We compute the following neuroretinal rim metrics to assess optic nerve health:

\begin{itemize}
    \item \textbf{NRR Area} ($A_{\text{NRR}}$): The area of the neuroretinal rim, calculated by subtracting the cup area from the disc area.

    \begin{equation}
    A_{\text{NRR}} = A_{\text{disc}} - A_{\text{cup}}
    \end{equation}

    \item \textbf{ISNT Quadrant Areas}: Measurements of the neuroretinal rim in the Inferior, Superior, Nasal, and Temporal quadrants to detect regional variations.
\end{itemize}

\subsubsection{Texture Features}
We calculate texture features using the Gray-Level Co-occurrence Matrix (GLCM) \cite{haralick1973textural}, which includes the following metrics:

\begin{itemize}
    \item \textbf{Contrast} ($C$)
    \item \textbf{Dissimilarity} ($D$)
    \item \textbf{Homogeneity} ($H$)
    \item \textbf{Energy} ($E$)
    \item \textbf{Correlation} ($R$)
    \item \textbf{Angular Second Moment} (ASM)
\end{itemize}

\subsubsection{Vessel Features}
We extract vessel features using the Frangi filter \cite{frangi1998multiscale} to enhance vascular structures:

\begin{equation}
V(x, y) = \max_{\sigma} \left( e^{-\frac{R_B^2}{2 \beta^2}} \left(1 - e^{-\frac{S^2}{2 c^2}} \right) \right)
\end{equation}
where $R_B$ is the blobness measure, $S$ is the second-order structureness, and $\sigma$ is the scale parameter.

% \subsection{Classification Model}

% \subsubsection{Network Architecture}
% We utilize a customized ConvNeXtTiny Model \cite{liu2022convnet}, initialized with ImageNet weights. The model architecture integrates the extracted features with the deep features from the CNN.

% \subsubsection{Model Details}
% The model consists of:

% \begin{itemize}
%     \item Input layer for standardized images.
%     \item ConvNeXtTiny backbone for feature extraction.
%     \item Global Average Pooling layer.
%     \item Concatenation with extracted features.
%     \item Fully connected layers with dropout for regularization.
%     \item Output layer with sigmoid activation for binary classification.
% \end{itemize}
% The model architecture utilizes a \textit{novel AGBO algorithm} that combines genetic algorithms with Bayesian optimization to balance exploration and exploitation capabilities, aiming to improve computational performance and achieve more efficient parameter tuning.

\subsection{Classification Model}

\subsubsection{Network Architecture}
We utilize a customized ConvNeXtTiny Model \cite{liu2022convnet}, initialized with ImageNet weights. The architecture employs a late fusion approach, integrating normalized manually extracted features with the deep features extracted by the ConvNeXtTiny backbone.

% \subsubsection{Model Details}
% The model consists of:
% \begin{itemize}
%     \item Input layer for standardized images
%     \item ConvNeXtTiny backbone for deep feature extraction
%     \item Global Average Pooling layer
%     \item Concatenation of normalized manual features with the pooled deep features (late fusion)
%     \item Fully connected layers with dropout for regularization
%     \item Output layer with sigmoid activation for binary classification
% \end{itemize}

% \subsubsection{Model Details}
% The model consists of:
% \begin{itemize}
%     \item Input layer for standardized images
%     \item ConvNeXtTiny backbone for deep feature extraction
%     \item Global Average Pooling layer
%     \item \textbf{Custom Classification Layers:}
%     \begin{itemize}
%         \item Sequential linear layers with ReLU activations and dropout for feature processing
%     \end{itemize}
%     \item Concatenation of normalized manual features with the processed deep features (late fusion)
%     \item Fully connected layers with dropout for regularization
%     \item Output layer with sigmoid activation for binary classification
% \end{itemize}

\subsubsection{Model Details}
The model consists of:
\begin{itemize}
    \item Input layer for standardized images
    \item ConvNeXtTiny backbone for deep feature extraction
    \item Global Average Pooling layer
    \item Custom classification layers with linear layers, ReLU activations, and dropout
    \item Concatenation of normalized manual features with the processed deep features (late fusion)
    \item Fully connected layers with linear layers, ReLU activations, and dropout
    \item Output layer with sigmoid activation
\end{itemize}

Concatenation occurs after the Custom classification layers, integrating manually extracted features with processed deep features before final classification layers, ensuring consistent scaling and enriching the feature set to capture complex patterns effectively.

This model leverages a \textit{novel AGBO algorithm} for hyperparameter tuning, combining genetic algorithms with Bayesian optimization to efficiently balance exploration and exploitation.

\begin{figure}[H]
    \centering
    \includegraphics[width=\linewidth]{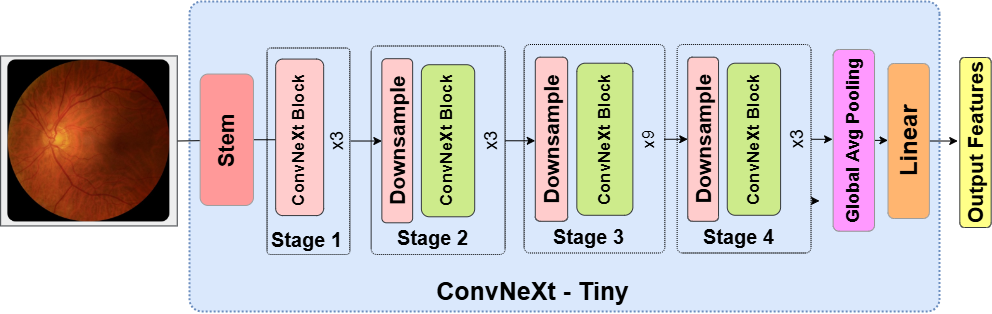}
    \caption{Architecture of ConvNeXtTiny-based classification network.}
    \label{fig:model_architecture}
\end{figure}

\subsubsection{Custom Loss Function}
We employ a custom loss function combining Focal Loss \cite{lin2017focal} and a false negative penalty to address class imbalance and emphasize correct detection of positive cases (glaucoma):

% \begin{equation}
% \mathcal{L}_{\text{total}} = \mathcal{L}_{\text{focal}} + \beta \mathcal{L}_{\text{FN}}
% \end{equation}

% \begin{equation}
% \mathcal{L}_{\text{focal}} = -\alpha (1 - \hat{y})^\gamma y \log(\hat{y}) - (1 - \alpha) \hat{y}^\gamma (1 - y) \log(1 - \hat{y})
% \end{equation}

% \begin{equation}
% \mathcal{L}_{\text{FN}} = y (1 - \hat{y})
% \end{equation}

\begin{align}
\mathcal{L}_{\text{total}} &= \mathcal{L}_{\text{focal}} + \beta \mathcal{L}_{\text{FN}} \\
\mathcal{L}_{\text{focal}} &= -\alpha (1 - \hat{y})^\gamma y \log(\hat{y}) \nonumber \\
&\quad - (1 - \alpha) \hat{y}^\gamma (1 - y) \log(1 - \hat{y}) \\
\mathcal{L}_{\text{FN}} &= y (1 - \hat{y})
\end{align}

where $\hat{y}$ is the predicted probability, $y$ is the ground truth label, $\alpha$ and $\gamma$ are hyperparameters of the focal loss, and $\beta$ controls the penalty for false negatives.

\subsection{Adaptive Genetic Bayesian Optimization (AGBO)}
We introduce the \textit{Adaptive Genetic Bayesian Optimization (AGBO)} algorithm, which integrates Bayesian Optimization with a Genetic Algorithm to effectively navigate the hyperparameter space.

\begin{figure}[H]
    \centering
    \includegraphics[width=\linewidth]{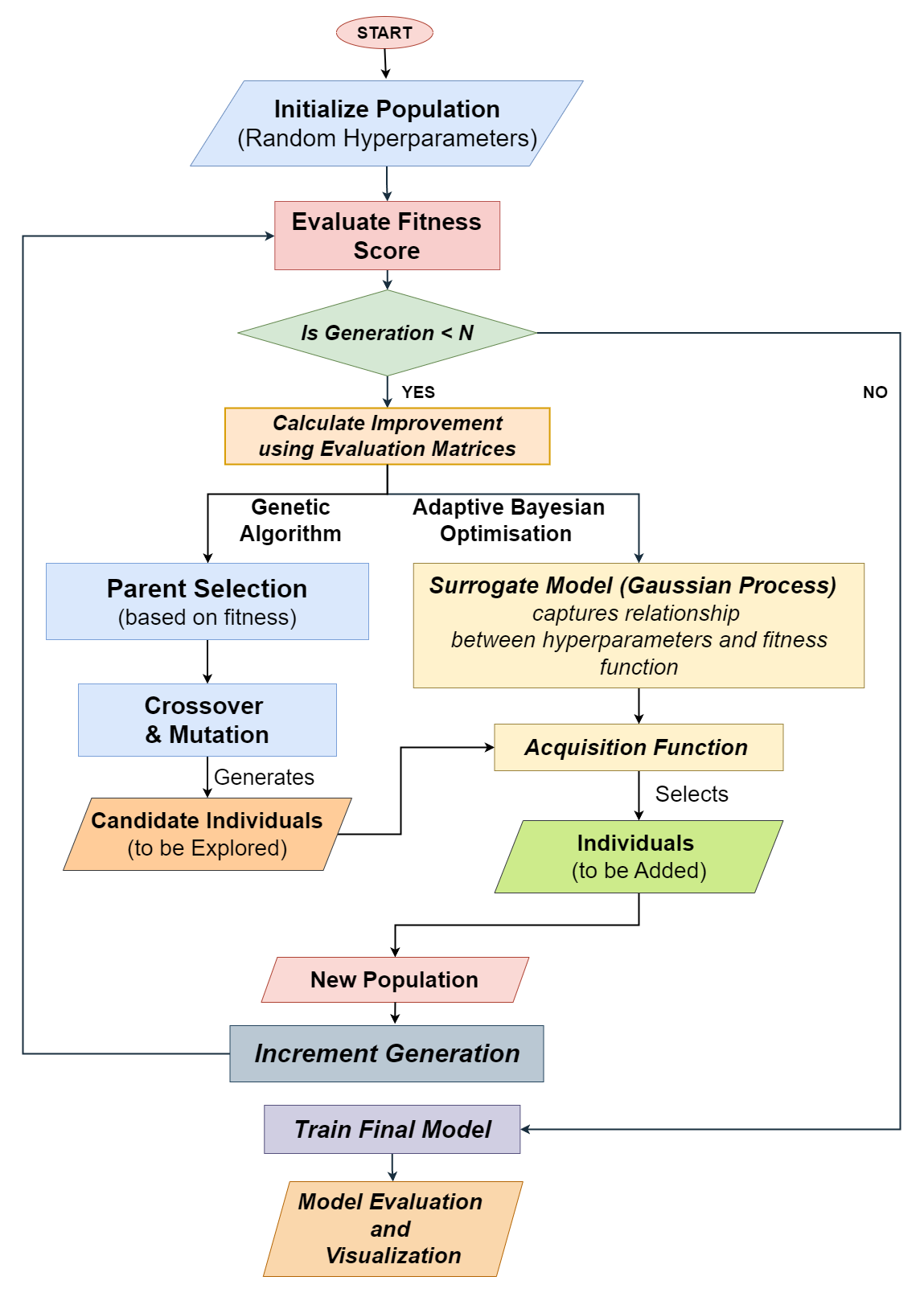}
    \caption{Flowchart of the proposed AGBO hyperparameter tuning process.}
    \label{fig:agbo_flowchart}
\end{figure}

AGBO begins by generating a diverse, random population of hyperparameters to ensure broad exploration of the search space. Each candidate's fitness is evaluated using the objective function, and genetic operators such as selection, crossover, and mutation  are applied iteratively to evolve the population, preserving diversity and enhancing the search for global optima.

In each iteration, a Gaussian Process (GP) surrogate model is trained on the historical evaluations to approximate the relationship between hyperparameters and their fitness scores. An acquisition function from Bayesian Optimization (e.g., Expected Improvement or Upper Confidence Bound) then uses the GP's predictions to rank candidates. Unlike traditional Bayesian Optimization, which typically proposes a small set of new points, AGBO leverages the Genetic Algorithm to create a larger candidate pool. From that pool, the acquisition function selects the single most promising candidate for actual evaluation, and the resulting \((\text{hyperparameters}, \text{fitness})\) pair is appended to the dataset \(\mathcal{D}\).

This hybrid approach balances exploration and exploitation by combining the genetic operators' ability to maintain population diversity with the Bayesian surrogate's capability for data-efficient guidance\cite{ABO}. As a result, AGBO converges more efficiently toward optimal hyperparameters, especially in larger or more complex search spaces. The process continues until a termination criterion, such as a maximum number of generations or minimal improvement in the objective, is reached.

\subsubsection{Algorithm Overview}
The proposed AGBO algorithm operates as follows:

\begin{algorithm}[htbp]
\caption{Adaptive Genetic Bayesian Optimization}
\KwIn{
  Hyperparameter space \( \mathcal{H} \),
  Objective function \( f(x) \),\\
  GP model with mean \( \mu(x) \), variance \( \sigma^2(x) \), kernel \( k(x, x') \),\\
  Acquisition function \( \alpha(x) \),
  GA parameters (\( p_c, p_m \)),\\
  Number of iterations \( T \)
}
\KwOut{Best solution \( x^* \) after \( T \) iterations}

\textbf{Initialize} dataset \( \mathcal{D} \) with \( N \) random hyperparameters from \( \mathcal{H} \)\;
\textbf{Initialize} GP model with \( \mathcal{D} \)\;

\For{\( t = 1 \) \KwTo \( T \)}{
    \textbf{Fit} GP model to \( \mathcal{D} \)\;

    \textbf{Initialize} candidate set \( \mathcal{C} \leftarrow \emptyset \)\;

    \textbf{Evolve population using a Genetic Algorithm on \( \mathcal{D} \)}:
    \Begin{
        \textbf{Selection} based on \( f(x) \)\;
        
        \textbf{Crossover} with probability \( p_c \), generating new candidates \( x \) and adding them to \( \mathcal{C} \)\;
        
        \textbf{Mutation} with probability \( p_m \) on each \( x \in \mathcal{C} \)\;
    }

    \textbf{Compute} acquisition function \( \alpha(x) \) for each \( x \in \mathcal{C} \)\;
    
    \textbf{Select} \( x_t \leftarrow \arg\max_{x \in \mathcal{C}} \alpha(x) \)\;
    
    \textbf{Evaluate} \( f(x_t) \)\;
    
    \textbf{Update} \( \mathcal{D} \leftarrow \mathcal{D} \cup \{(x_t, f(x_t))\} \)\;
}

\Return \( x^* \leftarrow \arg\max_{x \in \mathcal{D}} f(x) \)\;
\end{algorithm}

\subsubsection{Novelty of the Approach}
AGBO uniquely combines the population-based exploration of Genetic Algorithms with the principled, data-efficient guidance of Bayesian Optimization\cite{TEBO}. Traditional Bayesian Optimization might be slower in escaping local optima due to its limited sampling strategy, while Genetic Algorithms can maintain population diversity but often lack a fine-grained exploitation mechanism\cite{GAE}. By integrating these strengths, AGBO ensures robust exploration through evolutionary operators, while the GP surrogate and acquisition function guide exploitation by pinpointing promising regions of the hyperparameter space. Consequently, AGBO accelerates convergence and improves the likelihood of finding globally optimal solutions with a reduced computational budget. Moreover, the population dynamically adapts to insights gleaned from the Bayesian surrogate, further distinguishing AGBO from existing hybrid methods for complex hyperparameter tuning tasks \cite{10.1007/978-3-031-62269-4_42}.

\section{Experiments and Results}
In this section, we present the comprehensive evaluation of \textit{DeepEyeNet}, including an experimental setup, detailed analysis of classification performance, and an ablation study. \textit{DeepEyeNet} was benchmarked against existing models, and we analyzed the effects of various architectural choices through an ablation study to better understand the contribution of each component to the final model's performance.

\subsection{Experimental Setup}
Experiments were conducted using TensorFlow on a workstation with an NVIDIA A100 GPU (40 GB VRAM) and 80 GB system RAM. 
% Data augmentation techniques applied include random flips, rotations, zooms, and contrast adjustments to enhance model generalization. These augmentation techniques were essential for preventing overfitting and ensuring the model learned relevant features from the training data.

Data augmentation techniques applied include adding Gaussian noise, random horizontal and vertical flips, rotations, affine transformations with shearing and translation, slight brightness adjustments, and random resized cropping to slightly zoom and crop the images.
These augmentation methods enhance model generalization, prevent overfitting, and ensure the model learns relevant features from the data.

\subsection{Classification Performance}

To evaluate the efficacy of \textit{DeepEyeNet} in glaucoma detection, we compared its performance against several state-of-the-art models, including VGG16, ResNet50, EfficientNetB0, and DenseNet121.

% \textit{DeepEyeNet} was trained with optimized hyperparameters identified through the Adaptive Genetic Bayesian Optimization (AGBO) algorithm, as detailed in Table~\ref{table:deepeyenet_hparams}. Additionally, for a fair comparison, \textit{DeepEyeNet} was also trained using the same hyperparameters as the baseline models. The benchmark models were similarly trained with a batch size of 32, 20 epochs, and a learning rate of $1 \times 10^{-4}$, with their results averaged over two independent runs.

\textit{DeepEyeNet} was trained with optimized hyperparameters identified through our AGBO algorithm (see Table~\ref{table:deepeyenet_hparams}) and, for fair comparison, also with the baseline hyperparameters (batch size 32, 20 epochs, learning rate $1 \times 10^{-4}$). The benchmark models were trained under the same settings, and their results were averaged over two independent runs.

\begin{table}[h]
\caption{Best hyperparameters used to train the proposed \textit{DeepEyeNet} model (AGBO).}
\label{table:deepeyenet_hparams}
\centering
\begin{tabular}{|c|c|c|}
\hline
\textbf{Batch Size} & \textbf{Epochs} & \textbf{Learning Rate ($\eta$)} \\
\hline
8 & 80 & $1 \times 10^{-4}$ \\
\hline
\end{tabular}
\end{table}

\begin{table}[H]
\caption{Comparison with Other State-of-the-Art Models.}
\centering
\resizebox{\linewidth}{!}{%
\begin{tabular}{lccccc}
\toprule
\textbf{Method} & \textbf{Accuracy} & \textbf{Precision} & \textbf{Recall} & \textbf{F1-Score} & \textbf{AUC} \\
\midrule
\textbf{DeepEyeNet (AGBO)}       & \textbf{95.84\%} & \textbf{96.09\%} & \textbf{95.58\%} & \textbf{95.83\%} & \textbf{0.9848} \\
\textbf{DeepEyeNet (Baseline HP)}  &94.42\%           & 93.40\%          & 95.58\%           & 94.48\%          & 0.9817 \\
\textbf{VGG16}                     & 92.86\%          & 93.20\%          & 92.47\%          & 92.83\%           & 0.9731            \\
\textbf{ResNet50}                  & 92.28\%          & 91.89\%          & 92.73\%          & 92.31\%           & 0.9732            \\
\textbf{EfficientNetB0}            & 90.85\%          & 87.67\%          & 95.06\%          & 91.22\%           & 0.9712            \\
\textbf{DenseNet121}               & 93.12\%          & 94.52\%          & 91.56\%          & 93.00\%           & 0.9742            \\
\bottomrule
\end{tabular}%
}
\label{tab:classification}
\end{table}

As illustrated in Table~\ref{tab:classification}, \textit{DeepEyeNet (AGBO)} outperforms the baseline models, achieving significantly higher accuracy and robustness in distinguishing glaucomatous from healthy fundus images. When trained with baseline hyperparameters, \textit{DeepEyeNet} still maintains competitive performance, demonstrating both the robustness of the \textit{DeepEyeNet} architecture and the added benefit of fine-tuning with AGBO. This highlights the effectiveness of the proposed AGBO-optimized ConvNeXtTiny architecture in capturing relevant features for real-time glaucoma detection.

\begin{figure}[H]
    \centering
    \includegraphics[width=\linewidth]{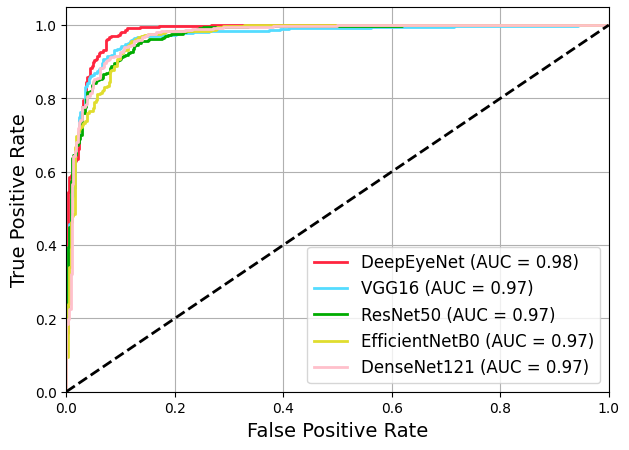}
    \caption{Comparison of ROC curves for \textit{DeepEyeNet} and other state-of-the-art models.}
    \label{fig:ROC_Comparison}
\end{figure}

% \begin{table}[H]
% \caption{Comparison with Other State-of-the-Art Models}
% \centering
% \resizebox{\linewidth}{!}{%
% \begin{tabular}{lccccc}
% \toprule
% \textbf{Method} & \textbf{Accuracy} & \textbf{Precision} & \textbf{Recall} & \textbf{F1-Score} & \textbf{AUC} \\
% \midrule
% \textbf{DeepEyeNet}                  & \textbf{95.84\%} & \textbf{96.09\%} & \textbf{95.58\%} & \textbf{95.83\%} & \textbf{0.9848} \\
% \textbf{VGG16}                        & 91.17\%          & 93.19\%          & 88.83\%          & 90.96\%           & 0.9701            \\
% \textbf{ResNet50}                     & 90.39\%          & 89.77\%          & 91.17\%          & 90.46\%           & 0.9702            \\
% \textbf{EfficientNetB0}               & 91.30\%          & 89.75\%          & 93.25\%          & 91.46\%           & 0.9682            \\
% \textbf{DenseNet121}                  & 91.30\%          & 90.15\%          & 92.73\%          & 91.42\%           & 0.9712            \\
% \bottomrule
% \end{tabular}%
% }
% \label{tab:classification}
% \end{table}

\subsubsection{Confusion Matrix}
The confusion matrix offers a comprehensive view of the \textit{DeepEyeNet} model's classification performance, showcasing a 95.84\% accuracy achieved. This highlights the model's strong capability in correctly classifying both glaucoma and non-glaucoma cases.

\begin{figure}[H]
    \centering
    \includegraphics[width=0.7\linewidth, height = 3.5cm]{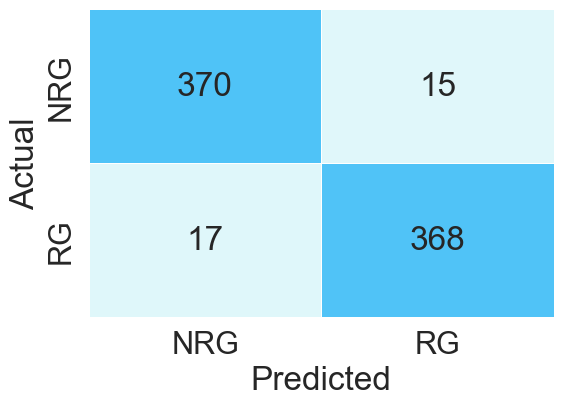}
    \caption{Confusion Matrix for DeepEyeNet}
    \label{fig:confusion_matrix}
\end{figure}

\subsection{Ablation Study}
We conducted an ablation study to assess the impact of different components on the \textit{DeepEyeNet} model's performance. Table~\ref{tab:ablation_comparison} presents the performance metrics for each configuration. The results highlight the importance of each component in the model's architecture. Removing image standardization and feature extraction led to a significant drop in accuracy (from 95.84\% to 93.12\%) and AUC-ROC (from 0.9848 to 0.9789), emphasizing their importance in preprocessing. Similarly, the absence of the proposed AGBO optimization reduced the accuracy to 94.42\% and AUC-ROC to 0.9817, demonstrating the critical role of effective hyperparameter tuning.

\begin{table}[H]
    \centering
    \caption{Performance Metrics for Different Model Configurations}
    \label{tab:ablation_comparison}
    \resizebox{\linewidth}{!}{%
    \begin{tabular}{lccccc}
        \hline
        \textbf{Method} & \textbf{Accuracy} & \textbf{Precision} & \textbf{Recall} & \textbf{F1-Score} & \textbf{AUC} \\
        \midrule
        \textbf{DeepEyeNet} & \textbf{95.84\%} & \textbf{96.09\%} & \textbf{95.58\%} & \textbf{95.83\%} & \textbf{0.9848} \\
        Without AGBO  & 94.42\% & 93.40\% & 95.58\% & 94.48\% & 0.9817 \\
        Without Image Std.\\\&Feat. Ext. & 93.12\% & 93.46\% & 92.73\% & 93.09\% & 0.9789 \\
        \hline
    \end{tabular}
    }
\end{table}

\begin{figure}[H]
    \centering
    \includegraphics[width=\linewidth]{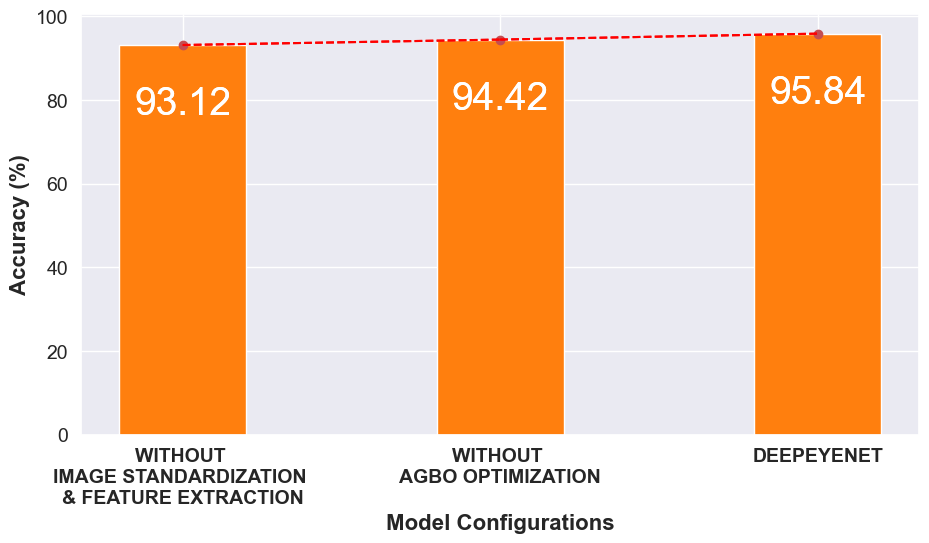}
    \caption{Ablation study results for accuracy of different model configurations.}
    \label{fig:ablation_results}
\end{figure}

The high AUC-ROC value of the \textit{DeepEyeNet} model (0.9848) reflects strong discriminatory power between classes, reinforcing the model's potential as a reliable tool for automated glaucoma screening.

\section{Discussion}

\subsection{Analysis of Results}
\textit{DeepEyeNet} demonstrates significant improvements over traditional methods. Advanced preprocessing, precise segmentation, comprehensive feature extraction, and hyperparameter optimization collectively enhance performance. The customized ConvNeXtTiny model, leveraging pre-trained weights, improves feature extraction, resulting in higher accuracy and AUC-ROC.

Exploration and exploitation are two fundamental components in hyperparameter optimization. The \textit{proposed AGBO algorithm} effectively balances these aspects. Exploration focuses on discovering new areas of the search space to identify potential hyperparameter configurations that have not yet been tested. In contrast, exploitation uses information gathered from previous evaluations to focus on areas that have demonstrated high performance. The genetic algorithm handles exploration\cite{CA1}, ensuring wide search space coverage\cite{CA2}, while the Gaussian Process model, integrated with Bayesian optimization\cite{HBO}, ensures effective exploitation by refining and focusing on regions with high potential \cite{snoek2012practical, franceschi2017forward}. The hybrid approach achieved faster convergence and a better trade-off between exploration and exploitation, avoiding premature convergence to suboptimal solutions.

In the hybrid framework, the genetic component prevents the search from stagnating, while the probabilistic modeling of Bayesian Optimization ensures that previously promising regions are exploited effectively \cite{young2015hyperparameter}. The success of this novel approach lies in its adaptability—able to explore extensively without losing the ability to focus on areas of high potential. The robustness, adaptability, and improved efficiency of \textit{DeepEyeNet} highlight its potential for practical deployment in clinical settings, where high performance and consistent reliability are crucial for early glaucoma screening and intervention.

\subsection{Key Insights}

There is significant potential to leverage AI in analyzing fundus images for developing automated glaucoma screening solutions, particularly benefiting economically disadvantaged regions. AI-assisted analysis of color fundus images can be advantageous in two main scenarios. Firstly, in non-portable, office-based settings, it can aid in diagnosis and prioritize patients for referral, ultimately helping to reduce unnecessary referrals and alleviate strain on healthcare systems. Secondly, it can be incorporated into portable devices used in under-resourced areas, allowing ophthalmic technicians or nurses to conduct screenings efficiently.

% Active research is focusing on two-step AI frameworks for glaucoma detection, where the first step involves automatic segmentation of optic cup and disc contours. This segmentation enhances the interpretability of AI models, providing transparency regarding why a particular classification was made, which is crucial for gaining trust in AI-assisted diagnosis.

Active research is focusing on two-step AI frameworks for glaucoma detection, where the first step involves automatic segmentation of optic cup and disc contours\cite{JD2}. This segmentation enhances the interpretability of AI models, providing transparency regarding why a particular classification was made, which is crucial for gaining trust in AI-assisted diagnosis. Additionally, the segmented optic disc and cup maps can be visualized and interpreted by medical professionals, thereby reinforcing the method’s clinical applicability and fostering greater confidence in its diagnostic decisions\cite{LS}.

The proposed \textbf{DeepEyeNet} framework integrates fundus image standardization, precise segmentation, feature extraction, and optimized classification using a customized ConvNeXtTiny model for glaucoma detection. This novel approach consolidates the previously two-step AI process into a single streamlined framework, enhancing efficiency. Additionally, we introduce Adaptive Genetic Bayesian Optimization (AGBO) algorithm for hyperparameter tuning, effectively balancing exploration and exploitation to achieve superior model performance. Thus, the proposed DeepEyeNet framework serves as a decision support tool, enabling collaboration between AI, clinicians, and patients to determine treatment options based on available resources.

% \section{Conclusion}
% We present \textit{DeepEyeNet}, a framework combining fundus image standardization, precise segmentation, feature extraction, and optimized classification using a customized ConvNeXtTiny model for glaucoma detection. Achieving 95.84\% accuracy, \textit{DeepEyeNet} outperforms other state-of-the-art models, showcasing its potential as an effective tool for automated glaucoma screening. The pre-trained ConvNeXtTiny enhances feature extraction, while the AGBO ensures optimal hyperparameter tuning, contributing to the high performance and reliability of \textit{DeepEyeNet} in early glaucoma diagnosis. The primary drawback is that the training process requires significant computational resources, which may limit its accessibility for institutions without high-performance hardware. Future work shall aim to optimize the computational efficiency and expand the framework to diagnose other ocular diseases, enhancing its accessibility for broader clinical use.

\section{Conclusion}
We present \textit{DeepEyeNet}, a framework that integrates fundus image standardization, precise segmentation, comprehensive feature extraction, and a ConvNeXtTiny-based classifier for accurate glaucoma detection. With a 95.84\% accuracy, \textit{DeepEyeNet} outperforms existing state-of-the-art models, demonstrating its promise in automated glaucoma screening. The pre-trained ConvNeXtTiny effectively captures subtle retinal features, while the Adaptive Genetic Bayesian Optimization (AGBO) algorithm balances exploration and exploitation for optimal hyperparameter tuning. Although the training process can be computationally intensive, especially for large populations in genetic algorithms, AGBO generally evaluates fewer CNN trainings overall, making it more time-efficient in extensive search spaces. Future work will focus on optimizing computational efficiency and extending this approach to other ocular diseases, thereby broadening its clinical impact. 
Additionally, we will experiment with multiple benchmark datasets to further validate our findings and enhance the generalizability of DeepEyeNet.

\bibliographystyle{IEEEtran}
\bibliography{references}

\end{document}